# Real-time 3D Reconstruction on Construction Site using Visual SLAM and UAV

Zhexiong SHANG[1] and Zhigang SHEN[2]

[1] Ph.D. Candidate, Durham School of Architectural Engineering and Construction, University of Nebraska-Lincoln, 113 Nebraska Hall, Lincoln, NE 68588-0500; e-mail: szx0112@gmail.com

[2] Associate Professor, Durham School of Architectural Engineering and Construction, University of Nebraska-Lincoln, 113 Nebraska Hall, Lincoln, NE 68588-0500; e-mail: shen@unl.edu

**ABSTRACT**

3D reconstruction can be used as a platform to monitor the performance of activities on construction site, such as construction progress monitoring, structure inspection and post-disaster rescue. Comparing to other sensors, RGB image has the advantages of low-cost, texture rich and easy to implement that has been used as the primary method for 3D reconstruction in construction industry. However, the image-based 3D reconstruction always requires extended time to acquire and/or to process the image data, which limits its application on time critical projects. Recent progress in Visual Simultaneous Localization and Mapping (SLAM) make it possible to reconstruct a 3D map of construction site in real-time. Integrated with Unmanned Aerial Vehicle (UAV), the obstacles areas that are inaccessible for the ground equipment can also be sensed. Despite these advantages of visual SLAM and UAV, until now, such technique has not been fully investigated on construction site. Therefore, the objective of this research is to present a pilot study of using visual SLAM and UAV for real-time construction site reconstruction. The system architecture and the experimental setup are introduced, and the preliminary results and the potential applications using Visual SLAM and UAV on construction site are discussed.

**INTRODUCTION**

In recent years, 3D reconstruction and mapping using unmanned aerial vehicle (UAV) has gained a lot of interest in civil and construction industry. Compare to ground vehicles, UAV are quicker, safer and more cost effective (Gheisari and Esmaeili). It can easily reach areas that are inaccessible to manned vehicles and undertake tasks that are dangerous to humans. Currently, the primary sensing technique equipped on small UAVs is photogrammetry. Compared to other remote sensors, such as LiDAR, radar and ultrasound, photogrammetry relies on optical sensor that has the advantages of texture rich, low-cost and light weighted, and can satisfy the required level of accuracy for most applications (Dai et al. 2012).  With onboard cameras, UAV can take aerial images during flight and produces 3D point clouds through image-based post-processing techniques (Liu et al. 2014). Recent applications showed the potentials of using UAV to monitor construction progress (Lin et al. 2015), survey civil infrastructures (Chan et al. 2015), and document historical site (Themistocleous et al.

2016). However, due to the heavy computational workload of image processing and computer vision algorithms, the model generation process is still slow (Nex and Remondino 2014), which limits its implementation on time critical applications.

Simultaneous localization and mapping (SLAM) is an advanced technique in robotics community which was originally designed for a mobile robot to consistently build a map of an unknown environment and simultaneously estimates its location in this map (Durrant-Whyte and Bailey 2006). When camera is used as the only exteroceptive sensor, such technique is called Visual SLAM or VSLAM (Artieda et al. 2009). Similar as photogrammetry, Visual SLAM has the advantages of rich visual data and low-cost. Previous applications of Visual SLAM and UAV mainly focus on indoor scanning and mapping (Michael et al. 2012; Scherer and Zell 2013), and only little or no effort has actively tested this technique at outdoor construction site (Ham et al. 2016). For many real-world applications in construction industry, timely data input play a key role of project success. Therefore, the objective of this study is to fill this knowledge gap by investigating the feasibility of using visual SLAM and UAV for real-time scene reconstruction and 3D mapping on construction site.

The following paragraphs are organized as follows: In section 2, related works of 3D reconstruction using optical sensors and UAV are summarized. In section 3, the general workflow of visual SLAM is briefly demonstrated. In section 4, the system architecture applied in this study, which includes both the hardware and software configuration, is introduced. In section 5, the experimental setup for outdoor environment using the proposed system is presented. In section 6, the authors evaluated the accuracy of SLAM generated model and presents the preliminary results of three potential applications using Visual SLAM and UAV on construction site. Finally, the conclusion, limitation and future study are discussed.

**LITERATURE REVIEW**

In civil and construction industry, most image-based 3D models are developed with post-processing techniques, which is a cascaded collection of image processing and computer vision algorithms (Lowe 2004; Triggs et al. 1999). With such technique, early investigators evaluated the availability of using UAV and onboard camera to survey historical site for landscape and heritage documentation (Remondino 2011; Remondino et al. 2011). 3D reconstruction on working construction site can be used to estimate excavation volume and track earthwork process (Siebert and Teizer 2014). The excavation volume was measured by generating digital elevation model (DEM) and the earthwork progress was tracked by monitoring the change of cut and fill area during earthmoving and compacting activities. Integrating the image generated 3D model with 4D BIM, the performance of UAV-enabled construction process monitoring can be increased, objects such as on-site occlusions can be identified and removed from camera point of view, and the ideal paths of UAV can be pre-determined (Lin et al. 2015). The image-based 3D model is also able to assist in condition assessment of in-service building and civil infrastructures. A recent application reconstructed an accurate model of a large-scale timber truss bridge using UAV and onboard camera (Khaloo et al. 2017). The model generated based on a total of 22 flights with more than 2000 high resolution images. By comparing with LiDAR, the result showed that although the noise floor of image-based 3D model is nearly three times higher, the density of points is much larger, which shows a better performance on structural detail representations. Other related applications of image-based post-processing 3D modeling using UAV includes subsidence surveying in abandoned mine areas (Suh and Choi 2017), terrian mapping (Stumpf et al. 2013), disaster site reconstruction (Ferworn et al. 2011) and forest investigation (Wallace et al. 2012; Zarco-Tejada et al. 2014).

Compare to the offline modeling applications, in construction industry, only limited efforts have been found on real time image-based 3D reconstruction using UAV. Previous study generated 3D point cloud of building façades by mounting a Kinect sensor on a hovering UAV (Roca et al. 2013). The Kinect range sensor provides both visual and depth image that can generated 3D colored point cloud in real time. However, due to the limitations of infrared sensors, the study fails to reconstruct objects with highly reflective materials or under extreme light conditions. In (Michael et al. 2012), a novel experiment was carried out to reconstruct the layout of a earthquake-damaged multistory building using both a ground robot and a small UAV. The ground robot equipped with an onboard rotating laser scanner provides feature-rich point cloud, the UAV attached both a laser scanner and a Kinect sensor take over the mapping tasks at areas where the ground robot cannot access. This study presented a potential of combining UAV and Visual SLAM for efficient 3D reconstruction and mapping on construction site.

**WORKFLOW OF VISUAL SLAM**

Although different algorithms of Visual SLAM existed, most studies follow the general workflow as feature extraction and matching, refinement of matching errors, loop closure detection and global map optimization (Fuentes-Pacheco et al. 2015). Compare to monocular camera, Visual SLAM with stereo camera provides both visual information and depth stream that increases the robustness of real-time mapping (Yousif et al. 2015). In this study, the authors implement a RGB-D graph-based SLAM approach introduced by (Labbe and Michaud 2013; Labbe and Michaud 2014) on a UAV. The approach applies appearance-based loop closure detection method, which uses bag of words, to associate the new images with the previous frames, and generate a global map with graph pose optimization. A three-layer memory management mechanism is designed to attenuate the accumulated memory usage for large-scale, long term, and multi-session mappings. Figure 1 shows the general workflow of using an RGB-D sensor and the graph-based SLAM approach for 3D reconstruction of an indoor office.

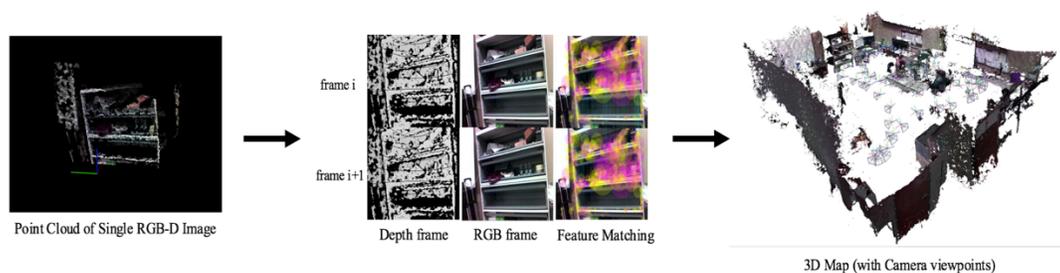

**Figure 1. Workflow of generating the 3D map of an indoor office using RGB-D camera and the graph-based SLAM approach**

## SYSTEM ARCHITECTURE

In this section, the hardware and software used for the implementation of Visual SLAM and UAV are introduced. The hardware consists of a UAV as an aerial platform, a stereoscopic vision system for visual and depth data feed and an onboard computer for SLAM computing. The DJI Matrices 600, which is a heavy loaded hexacopter UAV, is chosen as the aerial platform of this study. It can fly nearly 35 minutes with no payload and 15 minutes with full payloads (DJI 2016). The selected camera is the light weighted and portable RGB-D camera Intel Realsense R200, it contains a full HD color imaging sensor, an infrared laser projector and two registrated infrared IR sensor that can simultaneously provide both high resolution RGB video and depth stream (Intel 2016). The onboard SLAM computing is carried out by an NVIDIA Jetson TX1 processor. Robot Operating System (ROS) is installed in this computer to capture input video streams from R200, provide world coordinate system, send data to SLAM computing and output the reconstructed scene. The SLAM computing progress is executed by the open source software RTAB-Map (Real-Time Appearance-Based Mapping) introduced by (Labbe and Michaud 2013; Labbe and Michaud 2014). RTAB-Map is compatible with most commercial available RGB-D camera, such as Microsoft Kinect, Intel Realsense R200 and Intel Realsense ZR300 (Labbé 2013). Figure 2 presents the system diagram of the software and hardware configuration in this study.

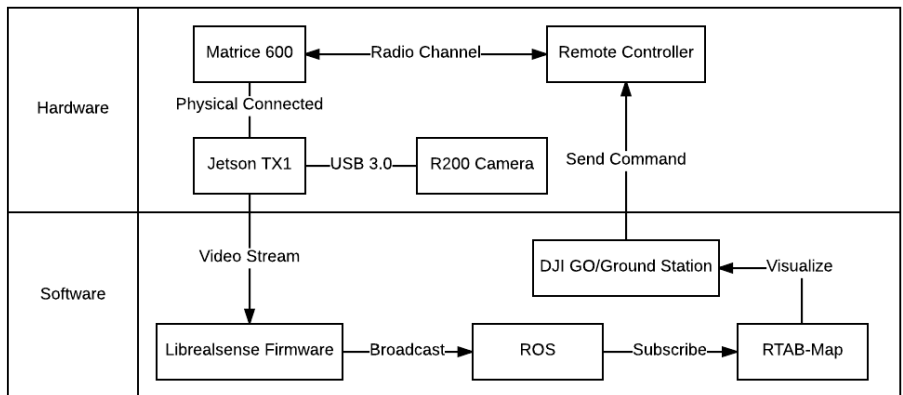

Figure 2. System diagram of the proposed system

## EXPERIMENTAL SETUP

In this study, the authors assume that UAV will always fly over the construction site, therefore, the R200 is attached at the bottom of the craft with face downwards to the ground. Due to the fact that RGB-D camera is sensitive to the illumination conditions, the parameters need to be well tuned in order to capture sufficient depth information under outdoor environment. For Visual SLAM, the best sensing distance of R200 ranges from 0.5 to 5 meters and the customized camera parameters for outdoor application are presented in Table. 1.

Table 1. R200 configuration for outdoor environment

| Number | Parameters | Values |
|---|---|---|
| 1 | Color_backlight_compensation | 0 |

| 2 | Color_enabled_auto_white_balance | 0 |
| 3 | R200_lr_auto_exposure_enabled | 1 |
| 4 | R200_lr_exposure | 23 |
| 5 | R200_lr_gain | 100 |
| 6 | R200_emitter_enabled | 1 |

Based on this condition, the UAV is designed at height between 4 to 5 meters above the ground both for safety and data reliability. The frame rate of RGB-D camera is set at 15 HZ with image resolution at 640 * 480. During the surveying process, the UAV flies autonomously with forward speed at 0.5 meters per second for sufficient feature detection and images alignment. An on-site operator manually controls the UAV travels to/return from the inspect region and avoids collisions with on-site objects. A DJI compatible gimbal camera is also attached on the aircraft for data comparison. The field set up and surveying process are shown in Figure 3.

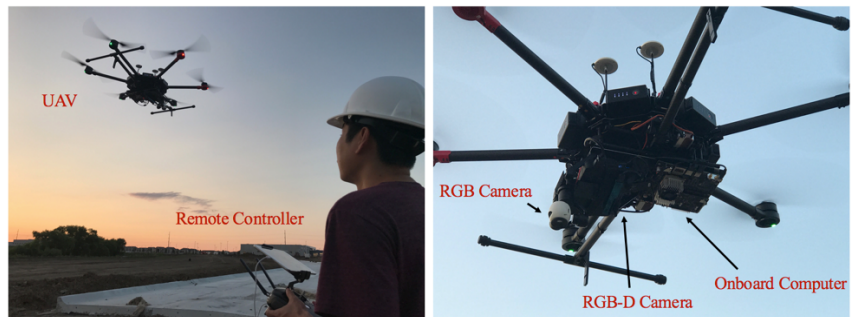

**Figure 3. Experimental Setup: (a) UAV and on-site operator; (b) the attached positions of RGB camera, RGB-D camera and onboard computer on the craft**

**FIELD APPLICATIONS**
In this section, the authors present the preliminary results of scene reconstruction on construction site with Visual SLAM and UAV. First, the model accuracy of Visual SLAM is evaluated by comparing it with post processing photogrammetry. Then, three applications that either uses Visual SLAM and UAV alone or combining it with other techniques on construction site are presented and discussed.

**Accuracy Evaluation**
To evaluate the performance of the scene reconstruction of Visual SLAM, a polygonal rubble trench foundation is selected as the measurement object. The result shows that although in many applications, the visual SLAM is considered as the real-time application of Structure from Motion (SfM), the point cloud density of the foundation generated with Visual SLAM (702089 points) is lower than the photogrammetry (1195326 points). Eight reference points (M0:M7 & M8:M15) located at each corner of the 3D point cloud model produced with both methods are manually selected, and the distances between the neighborhood points that forms the boundaries of the foundation are measured (Figure 4). The result shows that, comparing to photogrammetry, the average unit error of Visual SLAM is 3.3 centimeters without considering the measurement errors, which is acceptable for moderate scale mapping.

The big holes located at center of the Visual SLAM model shows the insufficient data capture caused by the limited sensing range of IR sensors, which also reduce the overall accuracy of Visual SLAM.

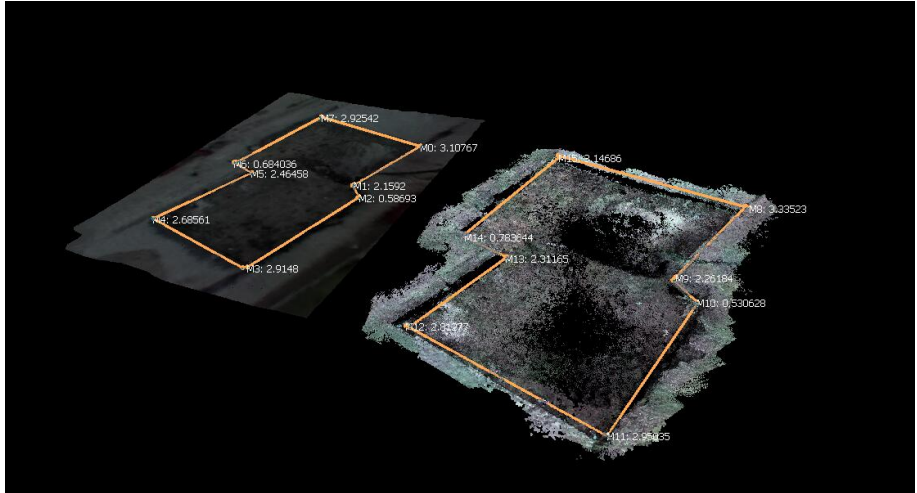

**Figure 4. Reference points and measured distances ($M_{i+1}$-$M_i$ in meters) of the point cloud models of a rubble trench foundation generated with Photogrammetry (left: M0:M7) and Visual SLAM (right: M8:M15)**

**Earthwork Measurement**

Previous efforts presented the applications of using the RGB camera and UAV to generate digital elevation model (DEM) of construction site and compute the earthmoving volumes (Siebert and Teizer 2014). Visual SLAM can generate similar models with nearly no post-processing time that largely shrinks the inspection interval from weeks or days to hours or minutes. Thus, more detailed construction progress can be monitored and analyzed. Figure 5 shows the point cloud models of the same trench foundation generated at different time stamps and the point cloud to point cloud (C2C) distance of earth volume changes are color coded.

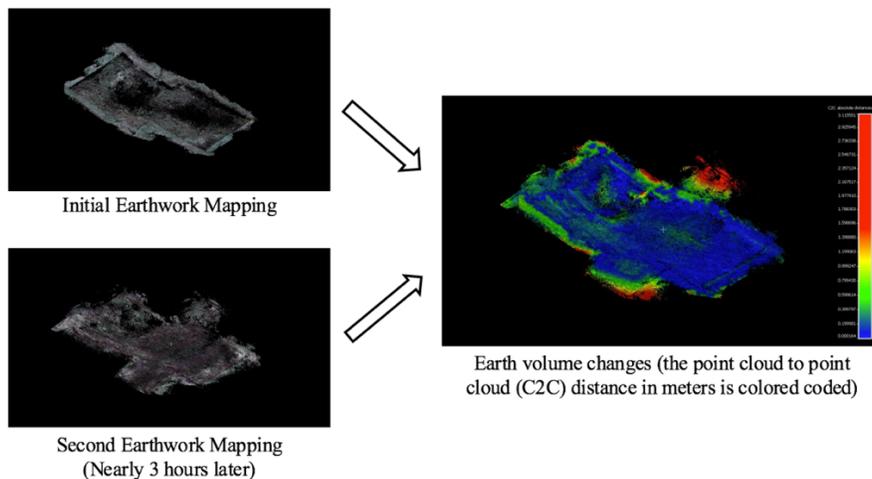

**Figure 5. Earth volume change measurement by comparing two SLAM models between three hours**

**Construction Progress Management**

It is known that building information modeling (BIM) is able to increase productivity and avoid conflicts on construction jobsite (Shang and Shen 2016). Integrating with 4D BIM, Visual SLAM is capable of monitoring detailed working progress on jobsite and efficiently detecting the task delay or changes. A prototype example of using BIM and visual SLAM for progress management of a pavement compaction task is presented in Figure 6. Figure 6 (a) shows the SLAM computed model of a jobsite that contains both site maps and working equipment. In figure 6 (b), the as-build BIM model is overlaid on the SLAM map to visualize the working progress, the different colors presented in the model shows the current state of pavement compaction progress.

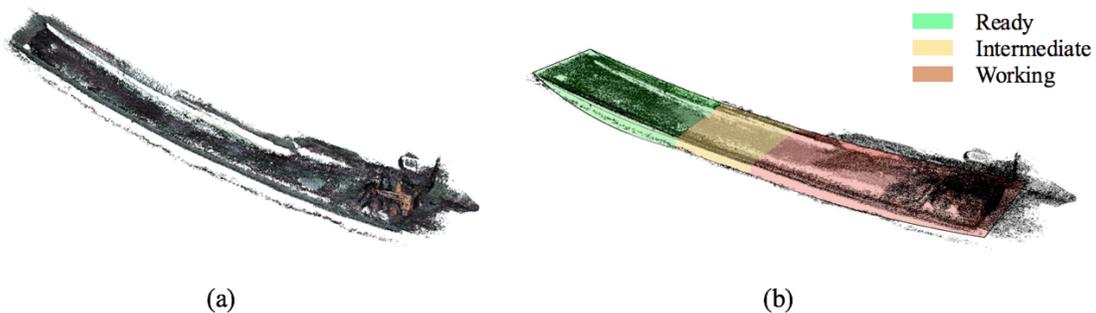

(a)                  (b)

**Figure 6. A site example that integrates SLAM model with 4D BIM for construction progress management**

**Site Asset Tracking**

A key technique of Visual SLAM is the visual odometry (VO), which is the process of estimating the camera's relative motion by analyzing a sequence of images. Combining with loop closure detection, this technique not only can produce the camera trajectory during the initial mapping progress, but also make it possible to locate a newly entered object, and at the same time, extends the map. Thus, site manager is able to track the real-time locations and trajectories of on-site objects (e.g. labors, equipment, materials) even under GPS-denied environment. The implementation of this technique can be classified into four major steps: 1) generate an initial site map with Visual SLAM. 2) share the map to other on-site equipment with either an online server or direct transmission. 3) attach an optical sensor on the moving object and trigger the VO and loop closure detection functions when it moves. 4) The trajectory of the objects will be identified in this map when the new captured images are associated with the global map. By enabling both the localization and mapping capabilities, the moving object can also extend the current map at areas where previous machines fail to access. An example of multi-session mapping and site object tracking using Visual SLAM on construction site is illustrated in Figure 7 below. Figure 7 (left) shows the initial map with trajectory of a sensing platform. In Figure 7 (right), the map is extended by different sensing platforms that the trajectory of each object is colored. Although this technique shows a good potential of on-site objects tracking and multi-session mapping with both aerial and ground platforms, the major limitation is the incremental memory usage for large area reconstruction. An experimental study shows that continuously mapping over 5000 frames will produce a map at size over 3.6 GB which decreased the performance of the onboard computer with a memory chip size around 4 GB.

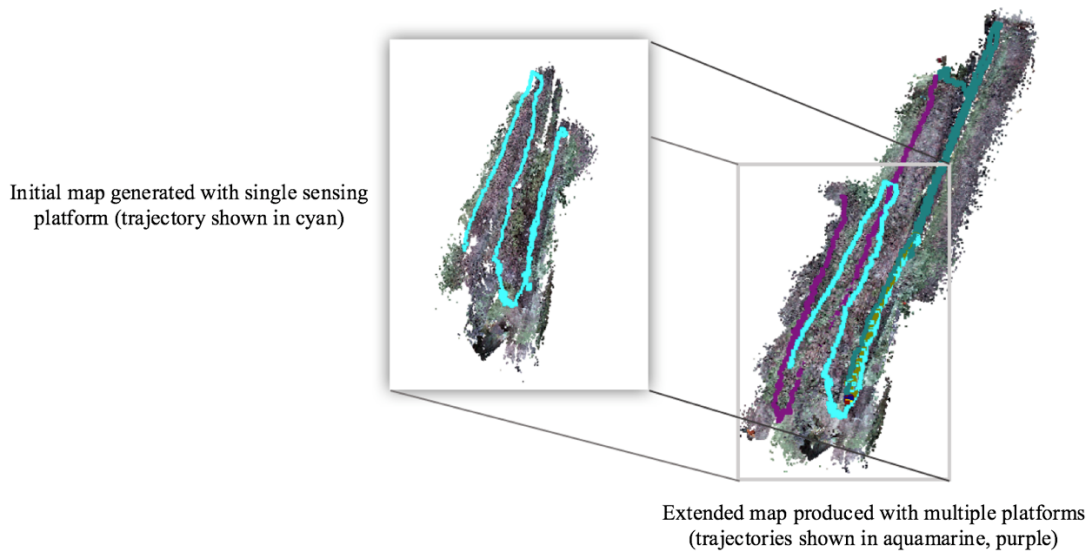

**Figure 7. Multi-session Mapping and Site Object Tracking using Visual SLAM**

**CONCLUSION**

This paper presents a preliminary study of using Visual SLAM and UAV at outdoor construction site. The hardware and software configuration, experimental setup, and the potential applications of using this technique on construction site are discussed. The preliminary results show that Visual SLAM and UAV can be a more efficient tool for 3D reconstruction than photogrammetry on time critical construction projects.

However, challenges still exist that limits the performance of current applications on large-scale and complex construction site, such as limited sensing range of stereo sensor, accumulated memory usage of onboard computer and the difficulty of maneuvering UAV at clutter environment. These problems can be potentially solved with more efficient SLAM computing algorithms, effective memory management mechanism and advanced flight strategy. Future study aims to overcome such limitations with sensor fusion technique and reactive UAV flight planning algorithms.